\documentclass[letterpaper, 10 pt, conference]{ieeeconf}  % Comment this line out if you need a4paper
\IEEEoverridecommandlockouts                              % This command is only needed if you want to use the \thanks command
\usepackage{amsmath}
\usepackage{amssymb}
\usepackage{graphicx}
\usepackage{multirow}
\usepackage[font={small}]{caption}
\usepackage{subcaption}
\usepackage{graphicx}

\title{\LARGE \bf
Monocular Simultaneous Localization and Mapping using Ground Textures
}

\author{Kyle M. Hart$^{1,2}$, Brendan Englot$^2$, Ryan P. O'Shea$^1$, John D. Kelly$^1$, David Martinez$^1$
    \thanks{$^1$Naval Air Warfare Center, Aircraft Division (NAWCAD), Lakehurst, NJ, 08733, USA,
        {\tt\small kyle.m.hart2.civ@us.navy.mil}
    }
    \thanks{$^2$Department of Mechanical Engineering, Stevens Institute of Technology, Hoboken, NJ, 07030, USA,
        {\tt\small benglot@stevens.edu}
    }
    \thanks{Distribution Statement A: Approved for public release; distribution is unlimited, as submitted under NAVAIR Public Release Authorization 2022-0586. The views expressed here are those of the authors and do not reflect the offical policy or position of the U.S. Navy, Deparment of Defense, or U.S Government.}
}
\begin{document}
\maketitle
\thispagestyle{empty}
\pagestyle{empty}

%%%%%%%%%%%%%%%%%%%%%%%%%%%%%%%%%%%%%%%%%%%%%%%%%%%%%%%%%%%
% Abstract 
%%%%%%%%%%%%%%%%%%%%%%%%%%%%%%%%%%%%%%%%%%%%%%%%%%%%%%%%%%%
\begin{abstract}
Recent work has shown impressive localization performance using only images of ground textures taken with a downward facing monocular camera. This provides a reliable navigation method that is robust to feature sparse environments and challenging lighting conditions. However, these localization methods require an existing map for comparison. Our work aims to relax the need for a map by introducing a full simultaneous localization and mapping (SLAM) system. By not requiring an existing map, setup times are minimized and the system is more robust to changing environments. This SLAM system uses a combination of several techniques to accomplish this. Image keypoints are identified and projected into the ground plane. These keypoints, visual bags of words, and several threshold parameters are then used to identify overlapping images and revisited areas. The system then uses robust M-estimators to estimate the transform between robot poses with overlapping images and revisited areas. These optimized estimates make up the map used for navigation. We show, through experimental data, that this system performs reliably on many ground textures, but not all.
\end{abstract}

%%%%%%%%%%%%%%%%%%%%%%%%%%%%%%%%%%%%%%%%%%%%%%%%%%%%%%%%%%%
% Introduction 
%%%%%%%%%%%%%%%%%%%%%%%%%%%%%%%%%%%%%%%%%%%%%%%%%%%%%%%%%%%
\section{Introduction}
\label{introduction}
When exploring unknown regions, robotic ground systems frequently rely on Simultaneous Localization and Mapping (SLAM) to map their surroundings and track their positions through the environment. While a wide range of sensors can be used, from lidar to vision, monocular cameras are a popular choice due to their low cost and rich information content.

These monocular SLAM systems frequently look out into the world for salient features to use for navigation. However, some environments, such as flat open spaces, lack enough features to reliably navigate. Additionally, cameras are sensitive to illumination changes in the environment, such as glare from a setting sun.

In these scenarios, the only consistent source of features comes from the surface the ground robot is traveling on. Recent work has shown that, despite its unstructured appearance, some ground textures are sufficiently distinct enough to successfully support localization \cite{schmid_features_2019, chen_streetmap_2018, kozak_ranger_2016}. These methods provide a reliable source of information, even in flat, open spaces or other feature sparse environments. Additionally, a downward facing camera is much more robust to illumination changes, since it has a limited, shielded field of view.

Here, we consider expanding upon ground texture localization methods by removing the requirement for an \textit{a priori} ground texture map. Without this need, an operator can save time on initial setup, since a complete map does not need to be created prior to operation. This allows exploration in previously unknown environments. Additionally, the system becomes more robust to changes in the environment that would differ from an \textit{a priori} map. This capability is accomplished through the introduction of a full SLAM system. In particular, our contributions in this systems paper are as follows:

\begin{itemize}
    \item To the best of our knowledge, development of the first online ground texture SLAM system using only a monocular camera.
    \item A unique algorithm within the ground texture domain that exploits the known depth of ground texture images when estimating the transform between overlapping images and identifying loop closures.
    \item Experimental results on a recent data set showing centimeter level accuracy on some textures and superior performance across changing textures, as well as consistent, accurate loop closure identification.
\end{itemize}

First, we will discuss some related works in Section~\ref{related-work} and formulate the target problem in Section~\ref{problem-description}. Then, Section~\ref{proposed-approach} will detail our approach to the target problem. Lastly, Section~\ref{experiments-and-results} will highlight supporting experimental results. The source code for this system is available at \textbf{https://github.com/Navy-RISE-Lab/ground-texture-slam}.
%%%%%%%%%%%%%%%%%%%%%%%%%%%%%%%%%%%%%%%%%%%%%%%%%%%%%%%%%%%
% Related work 
%%%%%%%%%%%%%%%%%%%%%%%%%%%%%%%%%%%%%%%%%%%%%%%%%%%%%%%%%%%
\section{Related Work}
\label{related-work}

Numerous SLAM systems incorporate monocular camera information. Some couple it with additional sensor information, such as inertial measurement units or lidar sensors \cite{shan_lvi-sam_2021, cheng_visual-laser-inertial_2021}. Others rely on just the camera as the only sensor. Our work aligns with this second class of systems.

Monocular camera only SLAM systems can be grouped into multiple categories distinguished both by the method used to compare images and by how the maps are stored. Image comparison generally falls into \textit{direct methods} or \textit{indirect methods}.  \textit{Direct methods} use the pixel intensities to compare images and include systems such as \cite{zubizarreta_direct_2020}. \textit{Indirect methods} use keypoints, which are salient points in the image identified through a variety of different algorithms, such as ORB and SIFT \cite{rublee_orb_2011, lowe_distinctive_2004}. Map storage typically includes \textit{dense} or \textit{sparse} maps. Dense maps store the entire image in the map, as in \cite{zhang_front-end_2021}. Sparse maps store a subset of information, such as keypoint values, as in \cite{zubizarreta_direct_2020}. The work described in this paper falls in the \textit{indirect-sparse} category.

Other \textit{indirect-sparse} monocular SLAM systems include the popular ORB-SLAM series \cite{mur-artal_orb-slam_2015, mur-artal_orb-slam2_2017, campos_orb-slam3_2021}. In these systems, keypoints are identified in images and their real-world locations are estimated to use for later localization and loop closure. Similar algorithms exist that look for domain-specific features, such as points and lines \cite{quan_monocular_2021, pumarola_pl-slam_2017}.

In many monocular SLAM systems, a major component of the algorithm attempts to perform accurate depth estimation. While images are information rich, the 2D nature loses depth information. To account for this, monocular SLAM systems frequently use either traditional methods for depth estimation \cite{civera_inverse_2008}, or use machine learning \cite{laina_deeper_2016,kim_revisiting_2021}.

For systems that use the ground texture for features, the depth estimation problem vanishes as all features are at the same depth. However, the current state of the art is limited to localization with an \textit{a priori} map of the environment. With these approaches, multiple images are taken of the ground such that the entire operating area can be found in at least one image. Then, each algorithm searches the map to find the closest matching images.

Most of these ground texture localization methods use keypoints and descriptors as in other monocular approaches \cite{schmid_features_2019}. Micro GPS uses a voting scheme where each keypoint votes for a closely matched image, then uses RANSAC to estimate a transform between images \cite{zhang_high-precision_2019}. \cite{fabian_schmid_ground_2020} uses a similar keypoint approach, but explores keypoint determination via both a uniform distribution across the image and random sampling without regard to image content. Lastly, \cite{zhang_learning_2018} uses a neural network to identify keypoints in images.

%%%%%%%%%%%%%%%%%%%%%%%%%%%%%%%%%%%%%%%%%%%%%%%%%%%%%%%%%%%
% Problem Description 
%%%%%%%%%%%%%%%%%%%%%%%%%%%%%%%%%%%%%%%%%%%%%%%%%%%%%%%%%%%
\section{Problem Description}
\label{problem-description}
\begin{figure}
    \centering
    \includegraphics[width=0.75\columnwidth]{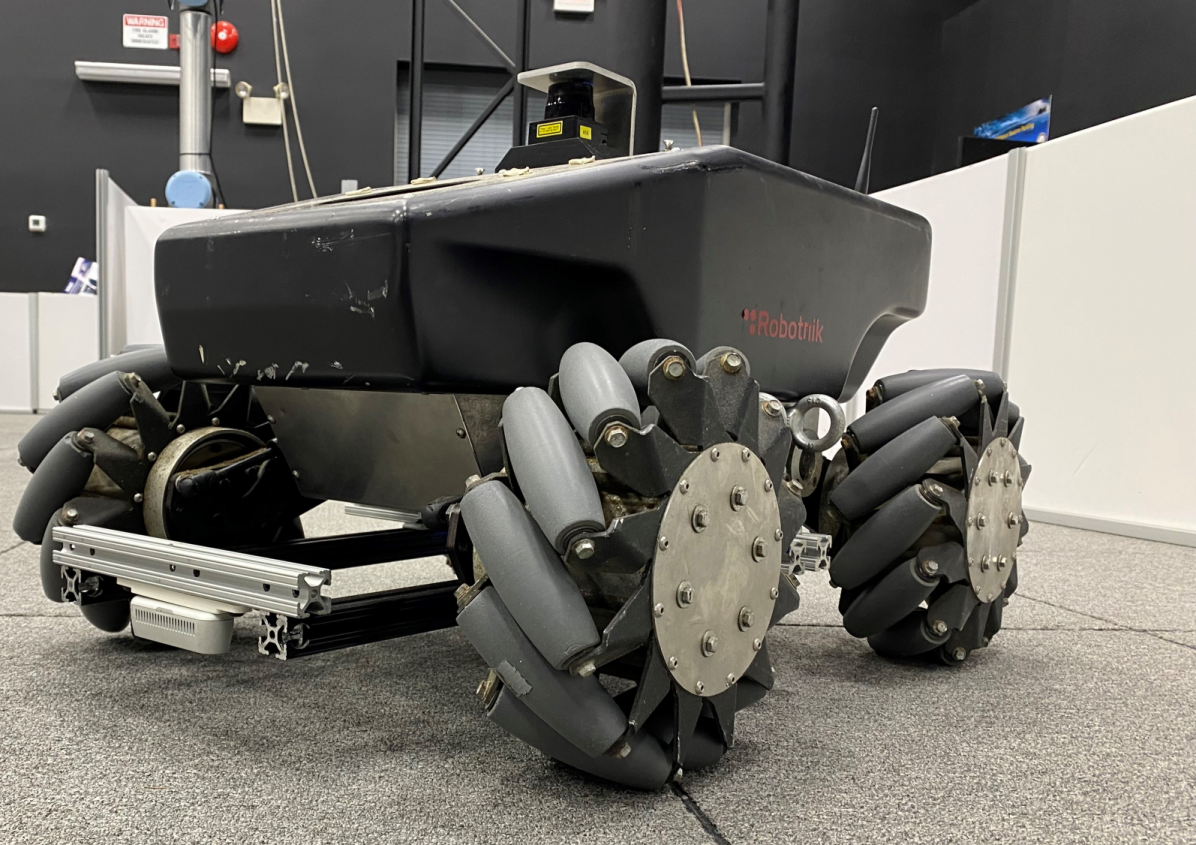}
    \caption{A downward facing camera setup configured by the authors. Note this is not the one used for gathering the experimental data considered in this paper, but is illustrative of a typical setup.}
    \label{fig:example-setup}
    \vspace{-4mm}
\end{figure}

Here, we consider a ground robot equipped only with a downward facing, calibrated, monocular camera with intrinsic matrix $\mathbf{K} \in \mathbb{R}^{3 \times 3}$ and known 3D pose relative to the robot's origin on the ground plane. This pose can be represented as the homogeneous matrix $\mathbf{T}_{RC} \in \mathbb{R}^{4 \times 4}$, which is used to transform data measured in the camera's frame of reference, $C$, into the robot's frame of reference, $R$. Fig.~\ref{fig:example-setup} shows an example setup.

The robot travels over a planar ground surface through several poses, $\Vec{x_t}$. These poses are defined in the ground plane as the robot's 2D pose as measured from the world or map frame, $W$:
\begin{equation}
    \Vec{x_t} = \begin{pmatrix} x_t\ y_t\ \theta_t \end{pmatrix}^\top
\end{equation}
and can be represented by the homogeneous transform $\mathbf{T}_{W\Vec{x_t}} \in \mathbb{R}^{3 \times 3}$. At each of these poses, the robot receives an observation, $\mathbf{Z}_t$, in the form of a distortion free image of the ground texture.

The goal is to develop an algorithm that can reliably estimate the robot's poses, $\Vec{x_t}$, for all $\mathit{t}$, using only the observations, $\mathbf{Z}_{0:t}$, the camera calibration matrix, $\mathbf{K}$, and the pose of the camera with respect to the robot, $\mathbf{T}_{RC}$.

Note that while odometry and inertial information are often available, this system specifically explores the ability to accomplish this goal without extra sensor information. Accomplishing this goal requires an approach that can estimate relative motion between successive images, and accurately detect previous sections of the terrain that have been revisited.
%%%%%%%%%%%%%%%%%%%%%%%%%%%%%%%%%%%%%%%%%%%%%%%%%%%%%%%%%%%
% Proposed Approach
%%%%%%%%%%%%%%%%%%%%%%%%%%%%%%%%%%%%%%%%%%%%%%%%%%%%%%%%%%%
\section{Proposed Approach}
\label{proposed-approach}
\begin{figure*}[tb]
    \centering
    \includegraphics[width=0.9\textwidth]{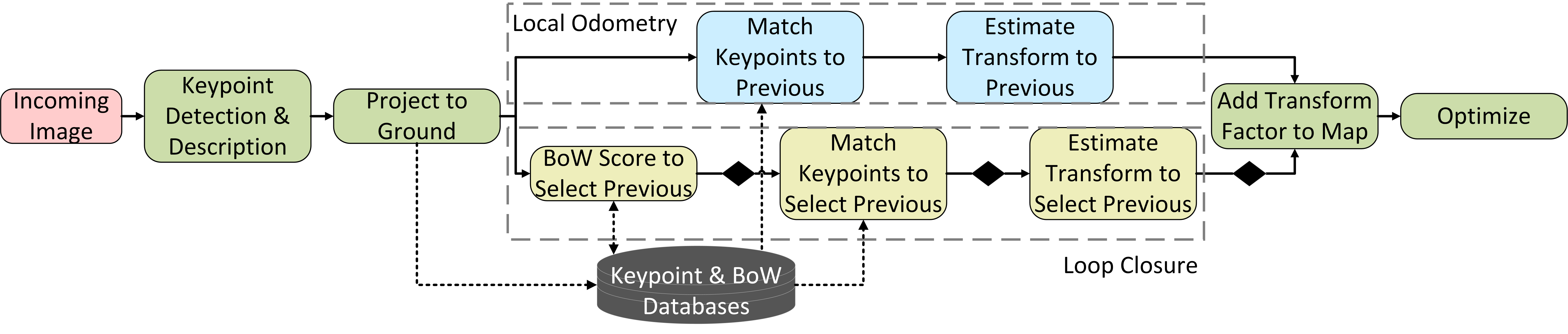}
    \caption{System structure for a single received image. There are two substructures: local odometry (blue boxes) that compares to the previous image; loop closure (yellow boxes) that compares to almost all previous images, with thresholding (indicated by the black diamonds). The remaining steps are either incoming data (red), general processing steps (green), or information stored between iterations (gray).}
    \label{fig:system-architecture}
    \vspace{-4mm}
\end{figure*}

We propose an algorithm that performs SLAM in three steps. Fig.~\ref{fig:system-architecture} shows an outline of the proposed approach. First, incoming images are processed. Then, successive pairs of images are used to estimate visual-only odometry. Then, loop closures are exploited to correct drift. Both the odometry and loop closure steps use identified keypoints from the image along with their associated descriptors. Additionally, both steps estimate the transform between pairs of images using keypoints projected into the ground plane and M-estimators, which are robust deterministic models. Unlike the local visual odometry, loop closure detection uses three metrics to determine if a candidate loop closure is valid. The transforms estimated from each step are inserted into a factor graph that represents the map. The overall process of the algorithm is described below.

\subsection{Image Processing}
\label{image-processing}
When an image is received, it is first processed to extract keypoints and their associated descriptions using the ORB algorithm \cite{rublee_orb_2011}. Then, the keypoints are converted from pixel points to ground points. By performing this conversion, the system can directly estimate the robot's transform in real-space instead of estimating the essential matrix and converting. In most 2D to 3D projection scenarios, this conversion is only to a scale factor of the depth. However, because the distance between the camera and ground plane is known, these points can be unambiguously projected. The points are first projected from pixel values to meters, as measured from the camera's frame of reference using the following equation:
\begin{equation}
    \mathbf{z_C} = \mathbf{K}^{-1} \cdot \mathbf{z_I} \times \mathit{d}.
\end{equation}
In this equation $\mathit{d}$ is the distance between the camera and the ground plane, as derived from $\mathbf{T}_{RC}$, and $\mathbf{z_I} \in \mathbb{R}^{3xN}$ is the collection of keypoints, in pixels, for this image, represented in homogeneous coordinates as follows:
\begin{equation}
    \mathbf{z_I} = \begin{bmatrix}
    \Vec{z}_I^{\;(0)} & \Vec{z}_I^{\;(1)} & \dots & \Vec{z}_I^{\;(N)}
    \end{bmatrix}
    =
    \begin{bmatrix}
    x_1 & x_2 & \dots & x_N \\
    y_1 & y_2 & \dots & y_N \\
    1 & 1 & \dots & 1
    \end{bmatrix}.
\end{equation}
The result is a collection of 3D vectors representing the point as measured from the camera's frame of reference, in meters (with some abuse of notation turning the homogeneous representation's 1 into the Z-component). This value can then be projected into the robot's frame of reference through conversion to a 3D homogeneous representation and a simple transformation:
\begin{equation}
    \mathbf{z_R} = \mathbf{T}_{RC} \cdot \mathbf{z_C}.
\end{equation}
From here, the Z components are dropped since the points and robot's pose are all equiplanar on the ground plane with a Z value of 0. This results in $\mathbf{z_R}$ as a collection of 2D points. After projection, the original pixel-valued keypoints are not saved. The projected keypoints and descriptors are retained for later use.

\subsection{Local Odometry}
\label{local-odometry}
To conduct local odometry, projected keypoints are matched to projected keypoints from the previous image. Then a transform is estimated between the images.

\subsubsection{Keypoint Matching}
\label{keypoint-matching}
Keypoint matching identifies corresponding keypoints that appear in two images using their descriptors. Our implementation uses the Fast Library for Approximate Nearest Neighbors (FLANN) method \cite{muja_fast_2009}. For each keypoint in the current image, a FLANN-based matching algorithm uses the keypoint descriptors to find two keypoints in the previous image that have the most similar descriptors. We use OpenCV's implementation \cite{bradski_opencv_2000}. This similarity is measured with a distance score, where lower scores indicate more similarity. If the two scores differ by a certain percentage, then the keypoints are considered to be a match. In other words, the keypoints match if they are significantly more similar than the next closest match. This ratio test was introduced in \cite{lowe_distinctive_2004}, and is applied as follows:
\begin{equation}
   \mathrm{match\ if\ } \mathit{f_0} <= \lambda \cdot \mathit{f_1}
\end{equation}
In the above inequality, $\mathit{f_0}$ and $\mathit{f_1}$ are the closest and second-closest distance scores, respectively, and $\lambda$ is the match threshold, which is often set between 0.5 and 0.7 in our method and 0.7 in the literature.

\subsubsection{Transform Estimation}
\label{transform-estimation}
Matched projected keypoints are then used in an M-Estimator factor graph to estimate the transform between them, using GTSAM's expression graph feature \cite{dellaert_borglabgtsam_2021}. The experiments described here use Huber, but others are available. This factor graph estimates the X, Y, and yaw components of the transform, $\mathbf{T}_{\Vec{x_j}\Vec{x_i}}$, which represents the transform between two of the robot's poses. The estimated transform is the one that best fits the below equation. Since the keypoints are projected onto the ground plane, as described in Section~\ref{image-processing}, this estimation occurs entirely in 2D real-space, offering increased efficiency over traditional 3D SLAM methods.
\begin{equation}
    \mathbf{z}_{R_j} = \mathbf{T}_{\Vec{x_j}\Vec{x_i}} \cdot \mathbf{z}_{R_i}
\end{equation}
Once estimated, the transform and associated covariance are added to the robot's SLAM factor graph as a factor between the current pose and previous pose. The system then proceeds to the loop closure identification step.

\subsection{Loop Closures}
To correct for drift, the system must correctly identify previously visited ground textures. The observations at all previously visited poses are possible candidates. Three threshold criteria are used: visual bag of words scores, the number of keypoint matches, and a covariance parameter.

\subsubsection{Visual Bag of Words}
The first threshold parameter is a Visual Bag of Words score. Using the technique and library described in \cite{galvez-lopez_bags_2012}, a database of previous image descriptors is built as new observations are received. To prevent redundant loop closures on adjacent observations, descriptors are not added to the database until a sufficient number of subsequent observations has been added. In other words, if the current observation just received is $\mathbf{Z_n}$, the descriptors from $\mathbf{Z_{n-k}}$ are added to the database.

The database is then queried with the current observation's descriptors to find matches. With the settings used in this work, returned scores from each previous observation in the database range from 0 to 1, with 1 being a perfect match. All results with a score above a certain threshold are considered candidate loop closures.

The vocabulary tree used to aid in descriptor matching in this step is assembled from the descriptors from a sample of images taken across all textures. As described in ORB-SLAM, this tree is general enough to work successfully on each sequence \cite{mur-artal_orb-slam_2015}.

\subsubsection{Number of Keypoint Matches}
Then, keypoint matching is performed, as in Section~\ref{keypoint-matching}. Any candidate loop closures with a number of matched keypoints less than the threshold are discarded.

\subsubsection{Covariance Parameter}
\label{covariance-parameter}
After discarding loop closure candidates that do not meet the previous threshold requirements, the remaining candidate loop closures have their transforms estimated as in Section~\ref{transform-estimation}. This procedure returns the estimated transform and a covariance matrix. The last threshold value is based on the covariance and is computed as the measure of maximum uncertainty of the estimate. The corresponding equation is as follows:
\begin{equation}
    \mathrm{score} = \log_{10}(\max(\mathrm{eigenvalues}(\Sigma))).
\end{equation}
In this equation, $\Sigma \in \mathbb{R}^{3 \times 3} $ is the covariance returned by the transform estimator.

This equation follows from the properties of eigenvalues as measures of magnitude along principal axes, therefore the maximum eigenvalue correlates to the maximum uncertainty. The logarithm provides a monotonic scaling factor to make tuning easier. Any potential loop closure greater than the threshold value is discarded.

To be considered a valid loop closure, a given candidate must meet all three threshold criteria. Notably, each criterion is checked as early in the loop closure algorithm as possible. This avoids the need for costly operations when available information could discard a candidate.

If a candidate loop closure meets all three criteria, it is then added to the SLAM factor graph as an additional factor between the two poses. The graph can then be optimized using the Levenberg-Marquardt algorithm in GTSAM and the system proceeds to the next image \cite{dellaert_borglabgtsam_2021}. This routine repeats throughout every image received during the robot's operation.

%%%%%%%%%%%%%%%%%%%%%%%%%%%%%%%%%%%%%%%%%%%%%%%%%%%%%%%%%%%
% Experiments and Results
%%%%%%%%%%%%%%%%%%%%%%%%%%%%%%%%%%%%%%%%%%%%%%%%%%%%%%%%%%%
\section{Experiments and Results}
\label{experiments-and-results}
\begin{figure}
    \centering
    \begin{subfigure}[b]{0.15\textwidth}
        \centering
        \includegraphics[width=\textwidth]{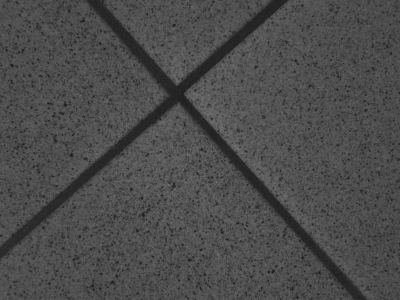}
        \caption{Bathroom Tiles}
    \end{subfigure}
    \hfill
    \begin{subfigure}[b]{0.15\textwidth}
        \centering
        \includegraphics[width=\textwidth]{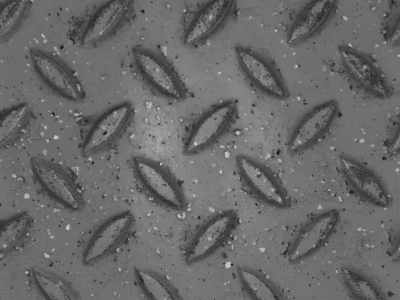}
        \caption{Checker Plate}
    \end{subfigure}
    \hfill
    \begin{subfigure}[b]{0.15\textwidth}
        \centering
        \includegraphics[width=\textwidth]{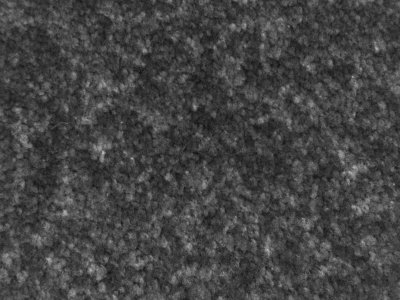}
        \caption{Doormat}
    \end{subfigure}
    \hfill
    \begin{subfigure}[b]{0.15\textwidth}
        \centering
        \includegraphics[width=\textwidth]{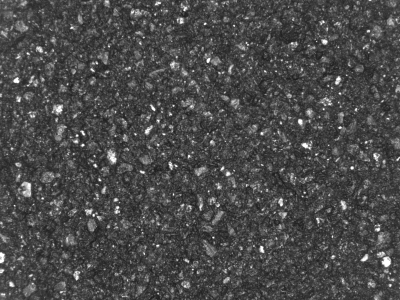}
        \caption{Footpath Asphalt}
    \end{subfigure}
    \hfill
    \begin{subfigure}[b]{0.15\textwidth}
        \centering
        \includegraphics[width=\textwidth]{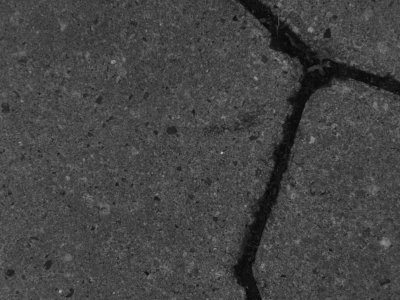}
        \caption{Parking Place}
    \end{subfigure}
    \hfill
    \begin{subfigure}[b]{0.15\textwidth}
        \centering
        \includegraphics[width=\textwidth]{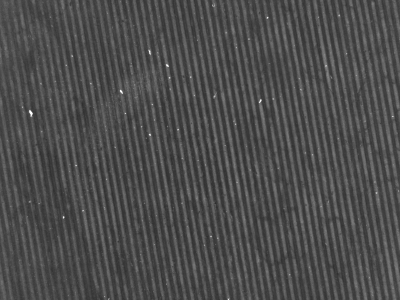}
        \caption{Ramp Rubber}
    \end{subfigure}
    \caption{Example textures from the dataset, licensed under CC BY-SA 4.0 \cite{schmid_hd_2022}. The dataset contains multiple texture environments. Each texture contains multiple sequences of observations. Each observation consists of an undistorted image and the associated ground truth at the time the image was captured.}
    \label{fig:example-texture}
    \vspace{-4mm}
\end{figure}

To validate this approach, results are conducted on the HD Ground Texture dataset \cite{schmid_hd_2022}. This dataset comprises multiple different environments with multiple paths through each environment captured with a downward facing camera setup as described in Section~\ref{problem-description}. The dataset also includes the ground truth poses at each image. Example images from the dataset are shown in Fig.~\ref{fig:example-texture}. This dataset only includes approximately planar surfaces, so the ability of this system to work on non-planar outdoor surfaces, like hills, remains untested.

\begin{figure}
    \centering
    \includegraphics[width=0.9\columnwidth]{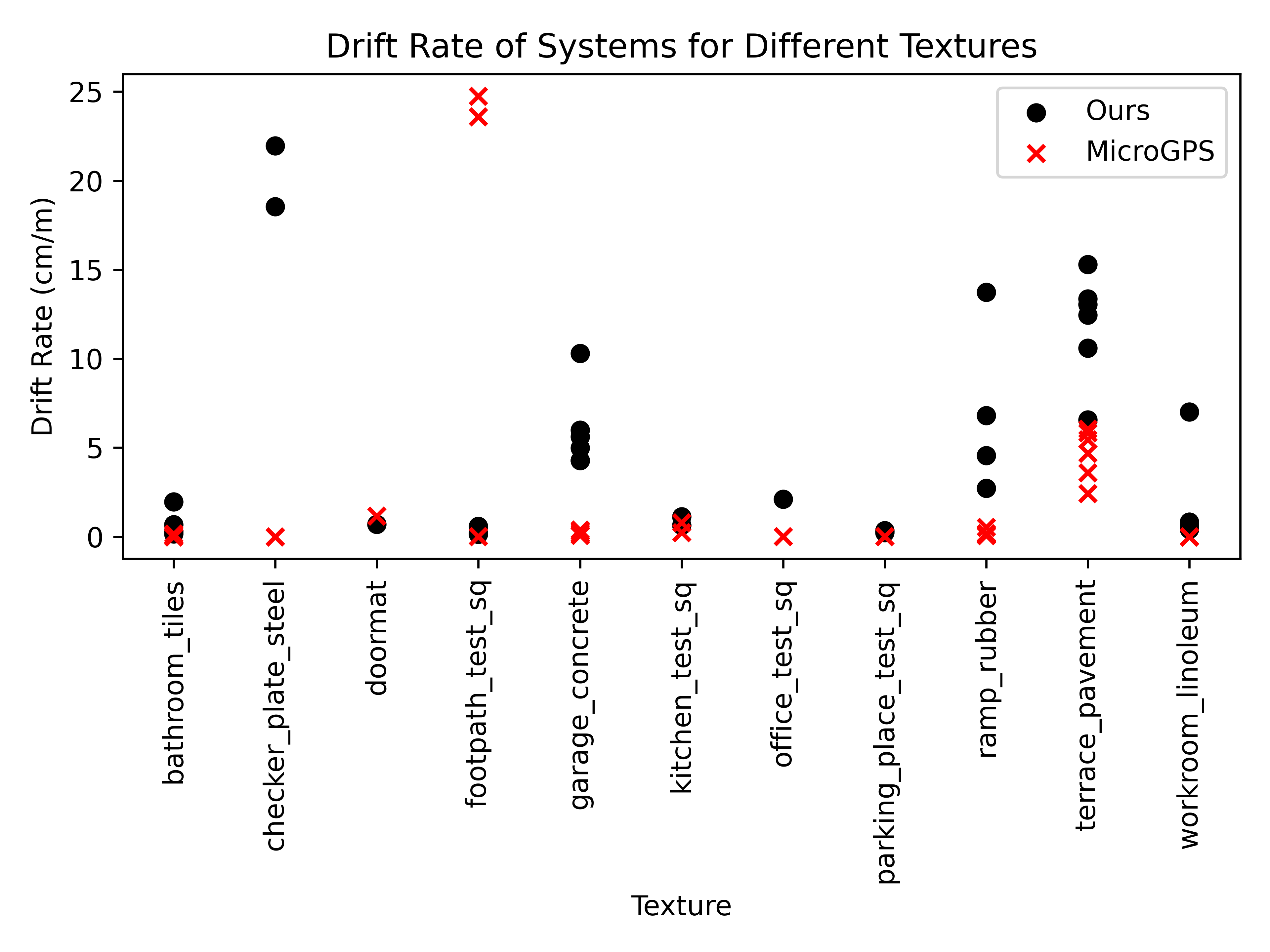}
    \caption{The translational mean absolute error for different paths and textures. Results are normalized by the total path length due to varying length. While Micro GPS often outperforms our system, it is a localization-only approach that needs an \textit{a priori} map.}
    \label{fig:accuracy-results}
    \vspace{-4mm}
\end{figure}

\begin{figure}
    \centering
    \includegraphics[width=0.9\columnwidth]{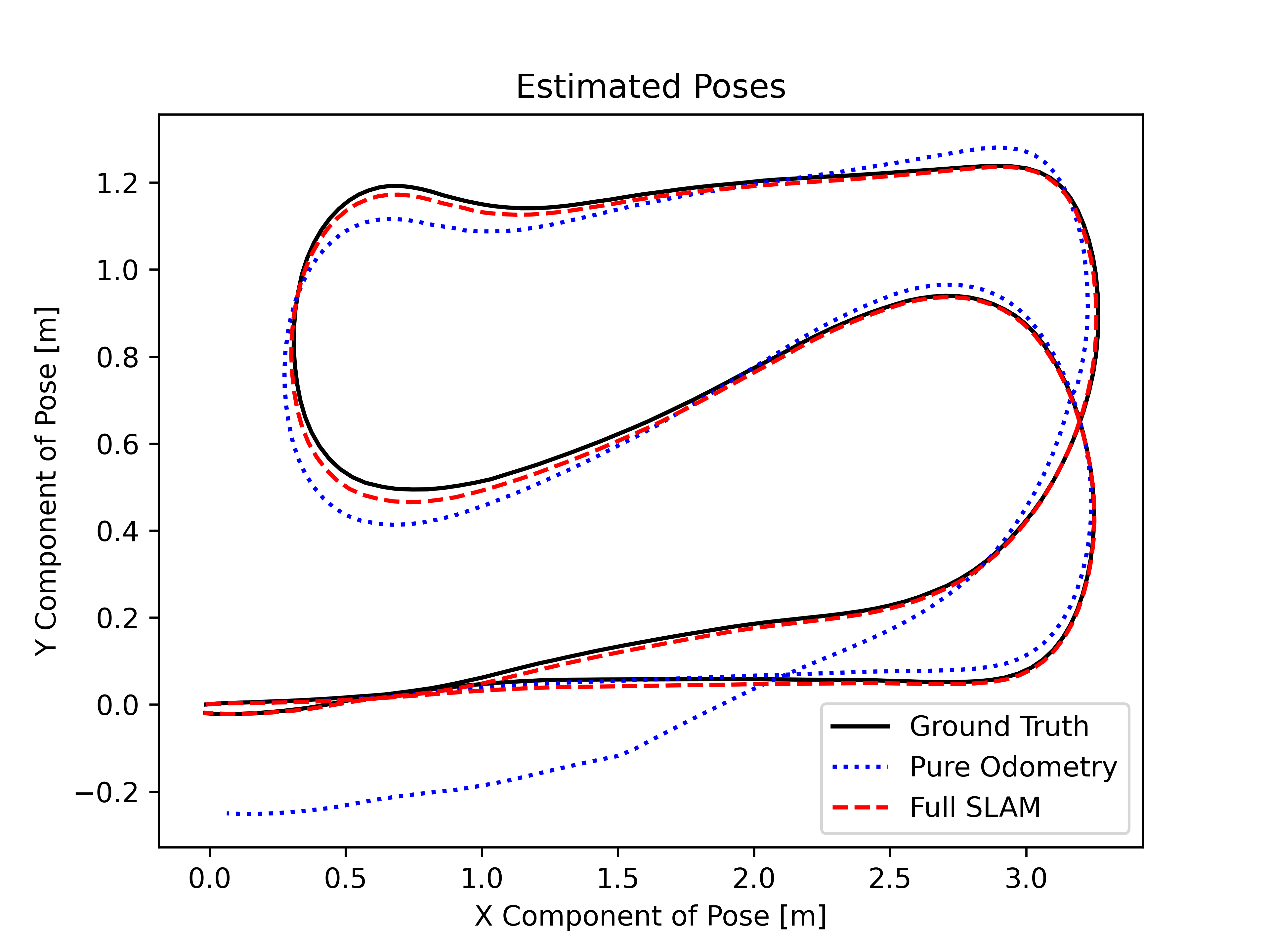}
    \caption{Example results for the \textit{Bathroom Tiles/test\_path1} texture and sequence. The red dashed line indicates the results of our SLAM system as described above. The blue dotted line indicates our system modified to perform without any loop closure corrections.}
    \label{fig:example-map}
    \vspace{-4mm}
\end{figure}

For testing, each image is loaded from file, then input into the SLAM system. After all images are input, the final estimated pose at each image is compared to ground truth. For comparison, Micro GPS is used to estimate poses for the same sequences \cite{zhang_high-precision_2019}. Micro GPS is a state-of-the-art localization system that estimates poses by comparison to a known map. Its default parameters are used for every texture. While it is expected that a localization system will outperform a SLAM system, this comparison establishes a baseline. The mean absolute translational error for multiple paths is shown in Fig.~\ref{fig:accuracy-results}, which is grouped by the type of ground texture. Due to varying path lengths, accuracy is normalized by total path length. Additionally, Fig.~\ref{fig:example-map} shows the results of one path over the one texture. This figure shows both our full SLAM solution and a variant of our solution without any loop closure, to indicate the effectiveness of the loop closure at correcting drift.

Our SLAM system shows reliable performance across six of the textures, with all reported error rates under 5 cm per meter traveled. Other textures show almost uniformly poor performance, indicating that characteristics of these textures make it difficult for this system to accurately perform. Textures with poor performance are typically caused by difficulty in matching successive images during the local odometry process described in Section~\ref{local-odometry} due to insufficient keypoint matches or inclusion of outliers. The loop closure stage is then unable to correct these errors\footnote{More detailed accuracy and loop closure results available in supplementary video and https://youtu.be/lJvTLQapsrQ}.

\begin{figure}[t!]
    \centering
    \begin{subfigure}[b]{0.9\columnwidth}
        \centering
        \includegraphics[width=\textwidth]{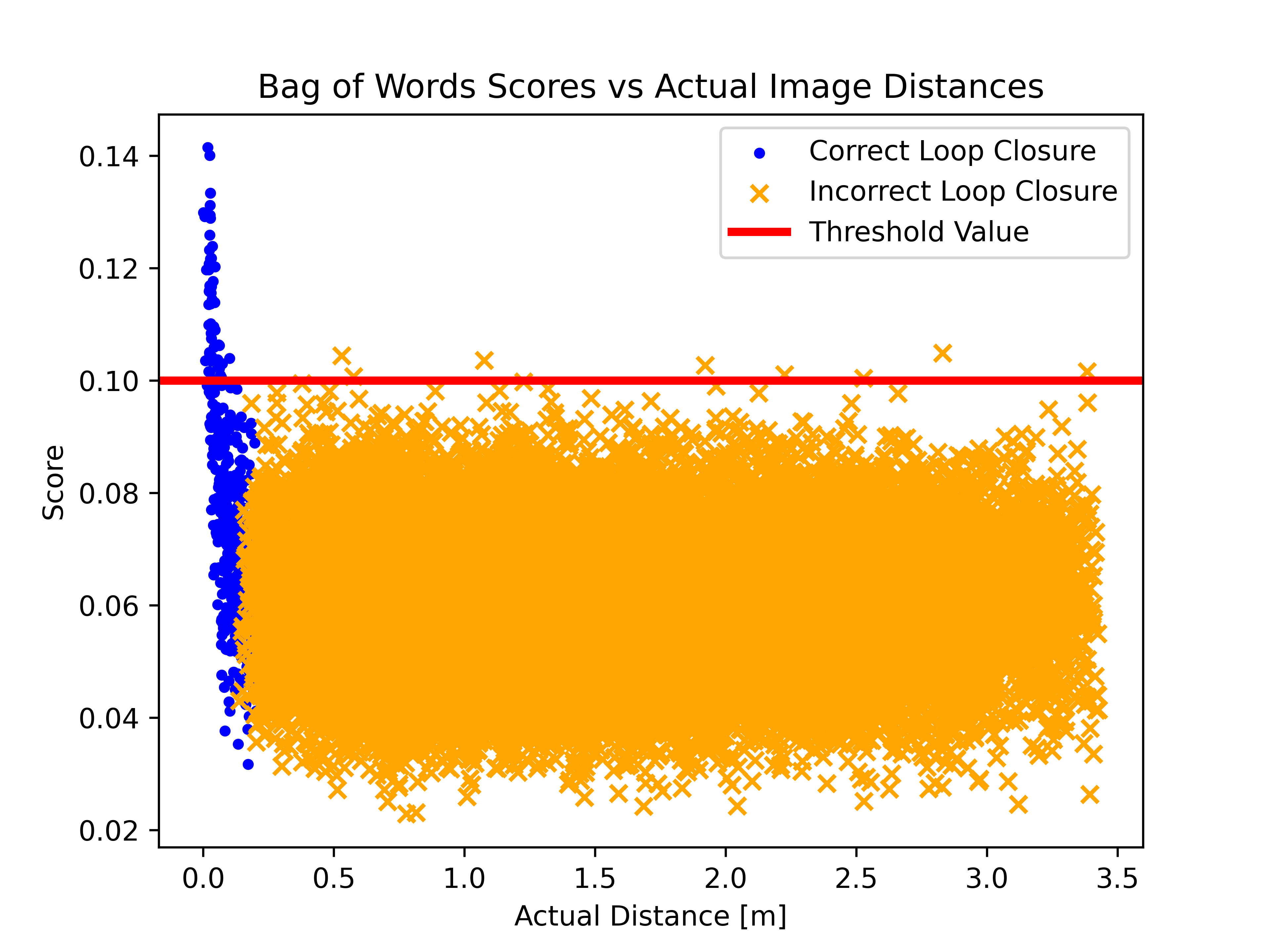}
        \caption{Bag of Words Scores}
        \label{fig:bow-plot}
    \end{subfigure}
    \hfill
    \begin{subfigure}[b]{0.9\columnwidth}
        \centering
        \includegraphics[width=\textwidth]{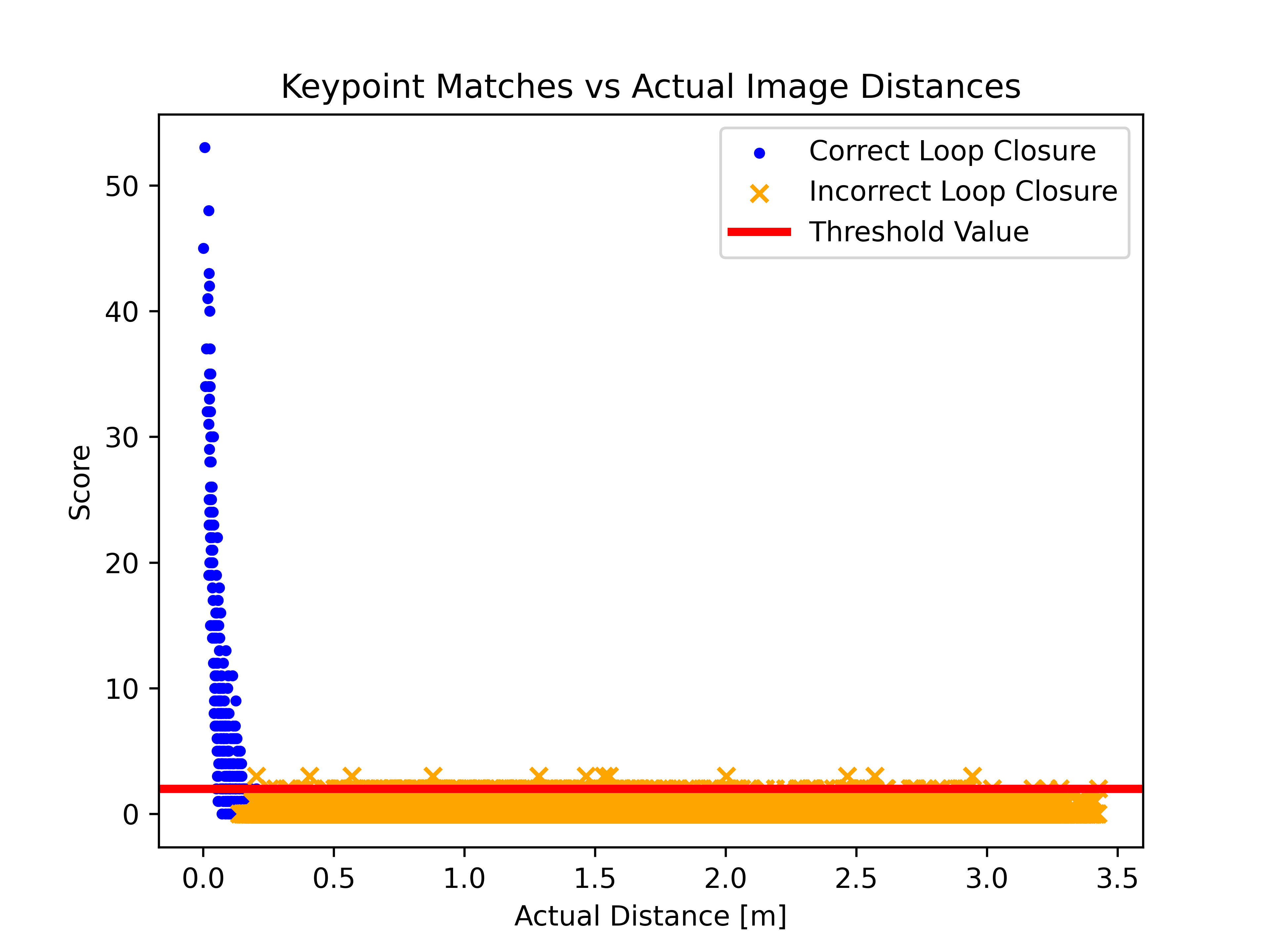}
        \caption{Keypoint Matches}
        \label{fig:keypoint-plot}
    \end{subfigure}
    \begin{subfigure}[b]{0.9\columnwidth}
        \centering
        \includegraphics[width=\textwidth]{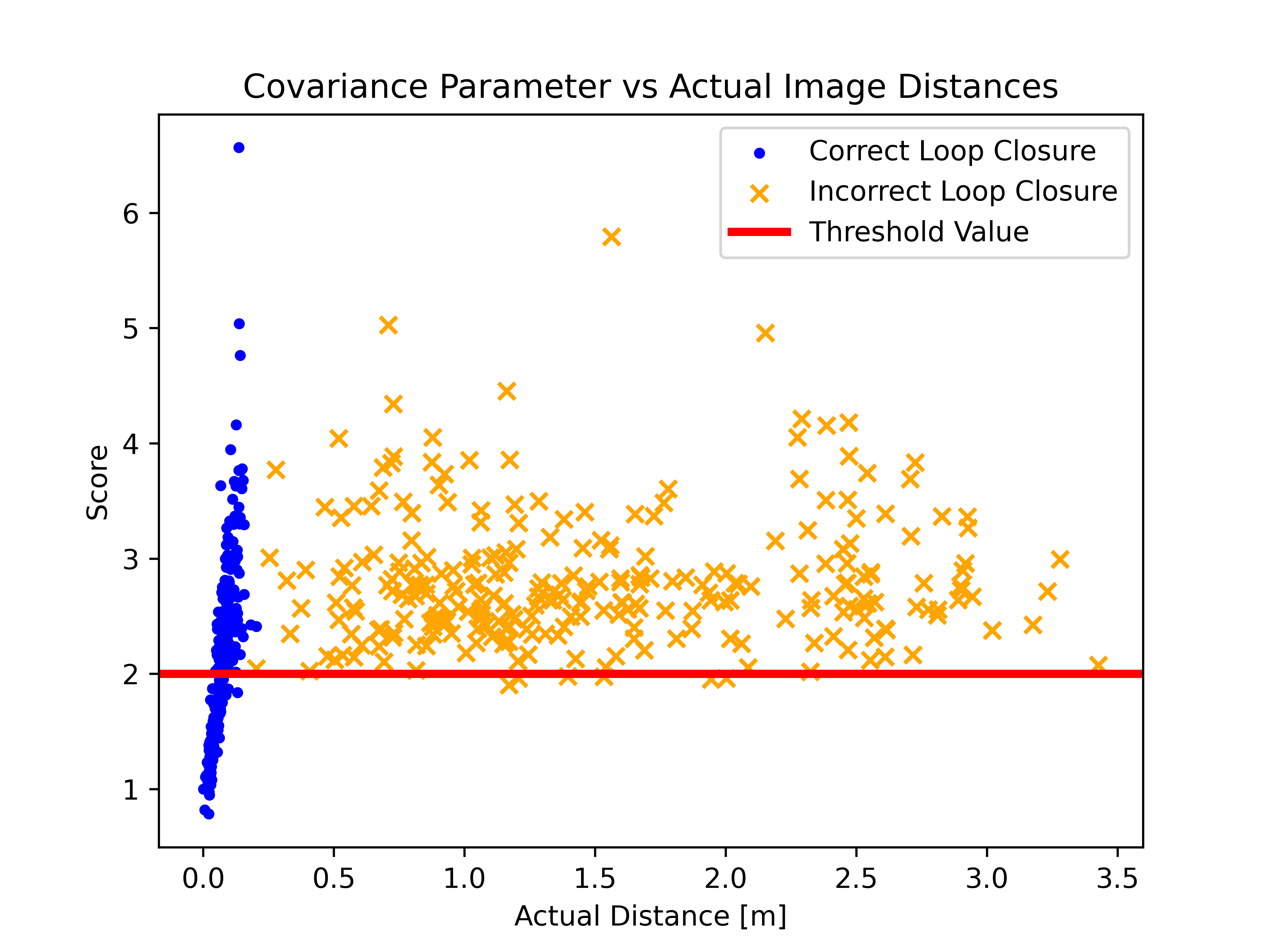}
        \caption{Covariance Scores}
        \label{fig:covariance-plot}
    \end{subfigure}
    \caption{The three threshold scores for pairs of images vs. the real world distance between those images. Only pairs of images with low actual distance can overlap and form valid loop closures, as shown. For (\subref{fig:bow-plot}) and (\subref{fig:keypoint-plot}) higher scores indicate better matches. For (\subref{fig:covariance-plot}) a lower score indicates a better match. Candidate loop closures that do not meet the threshold are removed from consideration.}
    \vspace{-4mm}
    \label{fig:score-plots}
\end{figure}

Regardless of overall system accuracy, the system shows good loop closure identification success rates, through use of the three threshold parameters. Fig.~\ref{fig:score-plots} plots threshold values for candidate loop closures for one of the ground textures in the HD Ground dataset \cite{schmid_hd_2022}. It also shows the actual real world distance between these images and identifies which are correct loop closures. Each threshold value, as indicated by the red lines, accurately removes a number of candidate loop closures. While not all correct loop closures are selected, very few, if any, incorrect loop closures are kept. These threshold values are determined by experimentation for each of the textures.

\begin{figure}[t]
    \centering
    \includegraphics[width=0.9\columnwidth]{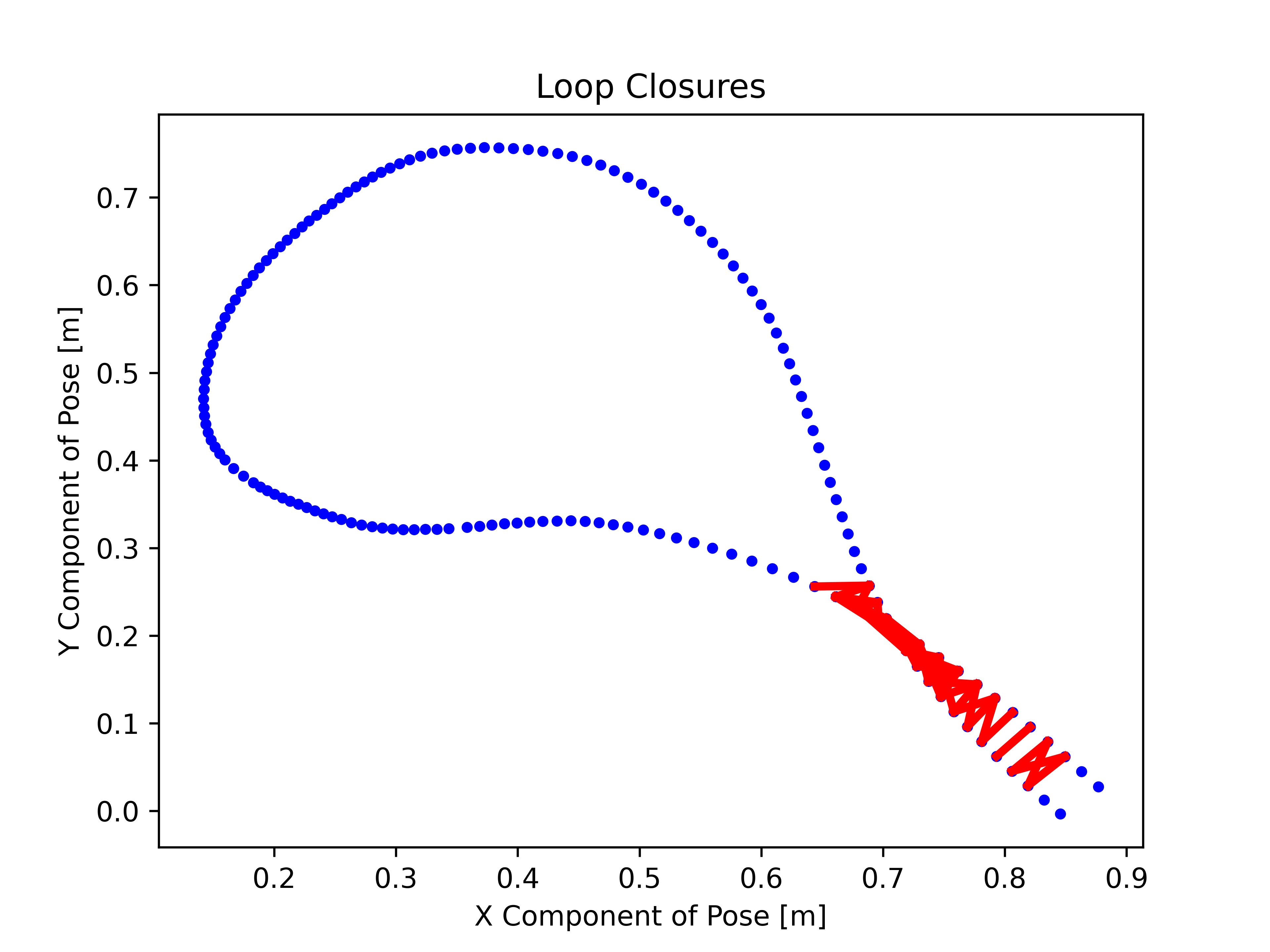}
    \caption{The loop closures, shown in red, identified during a representative trajectory, with robot poses shown in blue.}
    \label{fig:loop-closure}
    \vspace{-4mm}
\end{figure}

\begin{figure}[h]
    \centering
    \includegraphics[width=0.9\columnwidth]{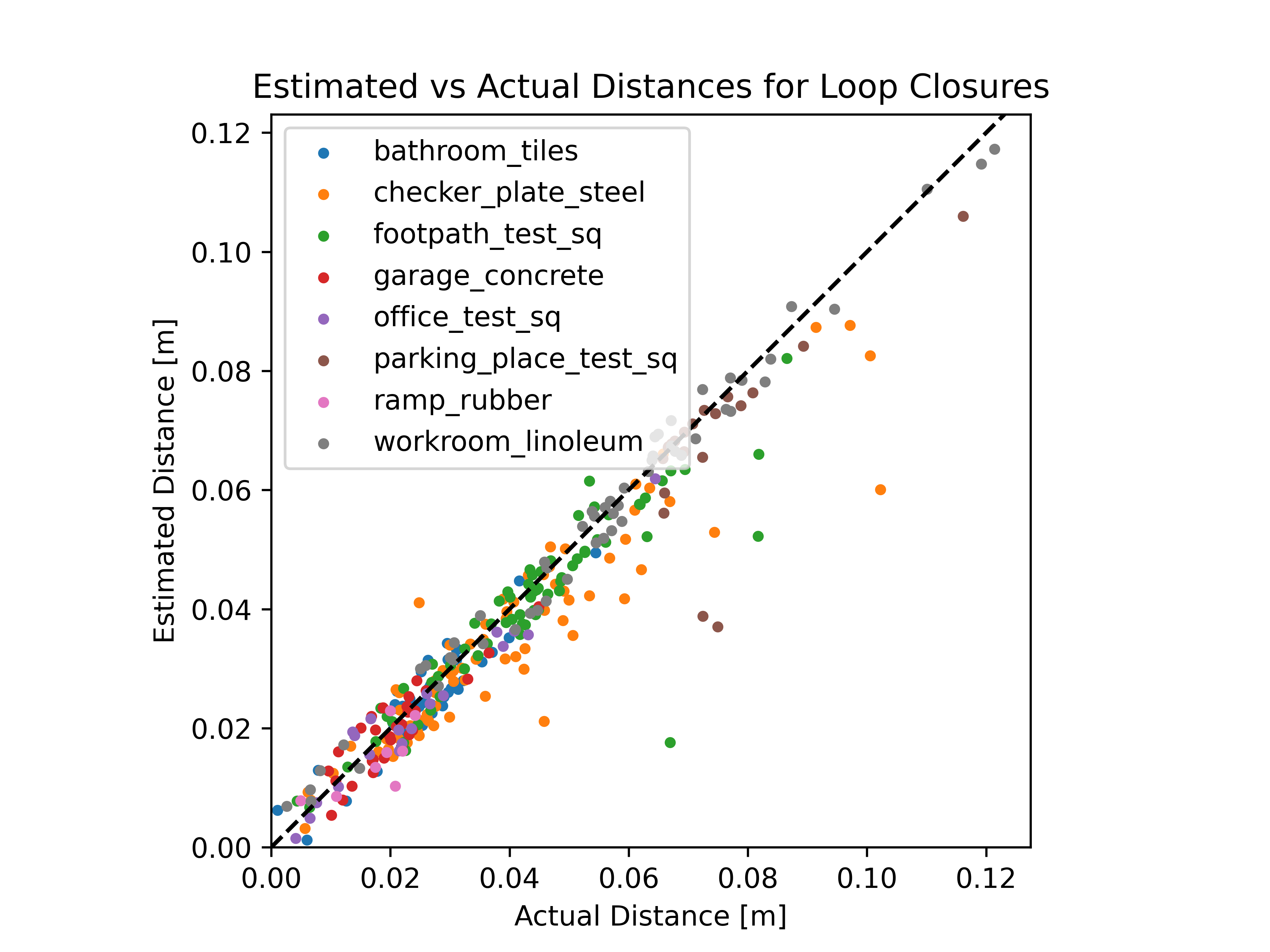}
    \caption{A plot of various estimated transforms between valid loop closure images pairs. One path from each texture is used for clarity. A y=x line indicating the ideal result is shown for reference.}
    \label{fig:loop-accuracy}
    \vspace{-4mm}
\end{figure}

Combining all three thresholds produces highly reliable loop closure selection. Fig.~\ref{fig:loop-closure} shows an example result, with red lines joining each identified loop closure. Note that each joins two nearby images without false positives. Additionally, Fig.~\ref{fig:loop-accuracy} shows a plot of the estimated distance of each selected loop closure versus the actual distance according to the ground truth for selected paths. In almost all cases, the estimated distance is very close to the actual, although some textures have outliers, including one which has been removed for clarity. This means that the proposed system is effective at both identifying loop closures and accurately measuring them for use in drift correction. Fig.~\ref{fig:example-map} illustrates an example of this by showing both our system and our system without any loop closure.

\begin{table*}[ht!]
    \centering
    \resizebox{0.99\textwidth}{!}{
    \begin{tabular}{|cc|cc|cc|}
        \hline
        \multirow{2}{*}{Texture} & \multirow{2}{*}{Path} & \multicolumn{2}{c|}{Normalized Translational MAE (cm/m)} & \multicolumn{2}{c|}{Rotational MAE (deg)} \\ \cline{3-6}
        & & Our SLAM & Micro GPS & Our SLAM & Micro GPS \\
        \hline
        Footpath Test Square & test\_path1 & 5.07 & \textbf{0.02} & 12.11 & \textbf{0.41} \\
        Footpath Test Square & test\_path2 & 1.96 & \textbf{0.01} & 5.49 & \textbf{0.35} \\
        Footpath Test Square & test\_path\_wet1 & \textbf{1.10} & 27.07 & \textbf{2.56} & 114.68 \\
        Footpath Test Square & test\_path\_wet2 & \textbf{2.15} & 24.42 & \textbf{5.78} & 107.29 \\
        Parking Place & regular\_test\_pp2\_201010 & \textbf{1.62} & 46.80 & \textbf{7.22} & 71.83 \\
        Parking Place & regular\_test\_pp2\_210225\_clean\_slightly\_wet & \textbf{6.60} & 24.73 & \textbf{19.14} & 30.21 \\
        Parking Place & regular\_test\_pp4\_210225\_slightly\_wet & \textbf{2.17} & 4.58 & 13.68 & \textbf{11.73} \\
        \hline
    \end{tabular}
    }
    \caption{A comparison of translational mean absolute error (MAE), normalized by overall path length, and rotational MAE between our SLAM system and Micro GPS \cite{zhang_high-precision_2019}. Sequences shown are highlighted because of changing appearance due to environmental effects.}
    \label{tab:weather-performance}
    \vspace{-4mm}
\end{table*}

Another result of note is the performance when the texture surface differs from when a map was created, such as after rain. Micro GPS compares captured images to an \textit{a priori} map for localization. Between capturing the map and capturing the images, rain changed the appearance of the ground on select textures, resulting in poor localization, as shown in Table~\ref{tab:weather-performance}. Since our SLAM system does not use an \textit{a priori} map, it is only comparing images of textures that are already wet, not comparing between wet and dry. This leads to more consistent performance. However, this may not hold true for very long duration SLAM sessions. Future work will explore those situations.

\begin{figure}[t]
    \centering
    \includegraphics[width=0.9\columnwidth]{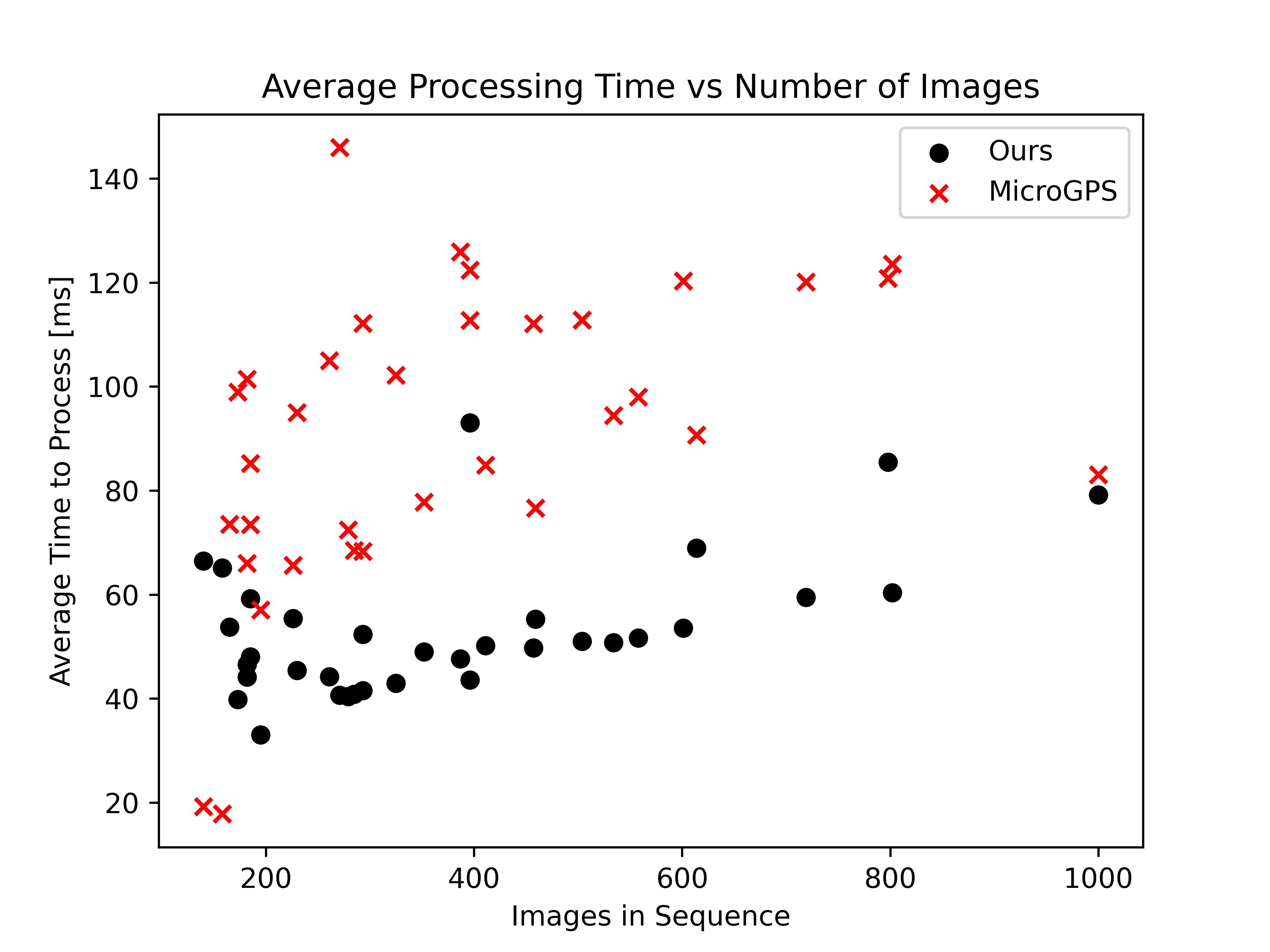}
    \caption{A plot of the processing speed of our full SLAM system and the Micro GPS localization system for various sequences.}
    \label{fig:time-results}
    \vspace{-4mm}
\end{figure}

In addition to accuracy, we also timed how long the system ran in each sequence. Fig.~\ref{fig:time-results} reports the results. It shows the number of images in a sequence and average processing time per image within the sequence, which approximates how fast images can be captured. These values are calculated by running a Docker container on a Windows 10 computer with an Intel i7 processor and 64 GB of RAM. The images are 1600 x 1200 pixels. Micro GPS values are also included for reference, but do not include the time required to collect and pre-process map data. This map-building time represents a significant duration beyond just runtime that our method does not require.

Notably, our system shows faster performance for the sequences tested due to the rapid candidate loop closure threshold evaluation. However, the data suggests longer sequences may prove unwieldy due to increasing times. Future work will look at map pruning to reduce time.

With both reliable accuracy and fast operating time, the proposed SLAM system provides an effective means of navigation over varied ground textures. Future work aims to realize a system that is more generalizable across a broad range of textures, is robust to environmental changes during multi-session SLAM, and is fast for long duration sessions.

%%%%%%%%%%%%%%%%%%%%%%%%%%%%%%%%%%%%%%%%%%%%%%%%%%%%%%%%%%%
% Conclusions 
%%%%%%%%%%%%%%%%%%%%%%%%%%%%%%%%%%%%%%%%%%%%%%%%%%%%%%%%%%%
\section{Conclusions}
\label{conclusions}
We have presented an innovative ground texture SLAM system that operates using only a calibrated, downward facing monocular camera. This system is the first to offer full online SLAM capabilities in the ground texture domain without the need for an existing map. When receiving a new image, it detects keypoints in the image and projects them onto the ground plane using known information about the camera position. It then uses robust M-estimator methods to estimate the ground plane 2D transforms experienced by the robot between pairs of images. Loop closures use three threshold values to identify revisited areas to improve overall accuracy. We have presented experimental results showing reliable performance on various ground textures and identified several that require further study to support. We have also shown accurate loop closure identification. When operating on acceptable textures, this system provides a means to rapidly set up robot navigation without the need to provide an \textit{a priori} map.

\section*{Acknowledgements}
This work was supported in part by NSF Grant IIS-1652064, and by the U.S. Naval Air Warfare Center - Aircraft Division's Naval Innovative Science and Engineering program. We would like to thank J. Fabian Schmid for giving us access to the HD Ground dataset.

%%%%%%%%%%%%%%%%%%%%%%%%%%%%%%%%%%%%%%%%%%%%%%%%%%%%%%%%%%%
% References
%%%%%%%%%%%%%%%%%%%%%%%%%%%%%%%%%%%%%%%%%%%%%%%%%%%%%%%%%%%
\bibliographystyle{IEEEtran}
\bibliography{IEEEabrv, root}

\begin{thebibliography}{10}
\providecommand{\url}[1]{#1}
\csname url@rmstyle\endcsname
\providecommand{\newblock}{\relax}
\providecommand{\bibinfo}[2]{#2}
\providecommand\BIBentrySTDinterwordspacing{\spaceskip=0pt\relax}
\providecommand\BIBentryALTinterwordstretchfactor{4}
\providecommand\BIBentryALTinterwordspacing{\spaceskip=\fontdimen2\font plus
\BIBentryALTinterwordstretchfactor\fontdimen3\font minus
  \fontdimen4\font\relax}
\providecommand\BIBforeignlanguage[2]{{%
\expandafter\ifx\csname l@#1\endcsname\relax
\typeout{** WARNING: IEEEtran.bst: No hyphenation pattern has been}%
\typeout{** loaded for the language `#1'. Using the pattern for}%
\typeout{** the default language instead.}%
\else
\language=\csname l@#1\endcsname
\fi
#2}}

\bibitem{schmid_features_2019}
J.~F. Schmid, S.~F. Simon, and R.~Mester, ``Features for ground texture based
  localization - a survey,'' in \emph{30th Brit. Mach. Vision Conf.}, Cardiff,
  UK, Sep 2019.

\bibitem{chen_streetmap_2018}
X.~Chen, A.~S. Vempati, and P.~Beardsley, ``{StreetMap} - mapping and
  localization on ground planes using a downward facing camera,'' in
  \emph{Proc. IEEE/RSJ Int. Conf. Intell. Robots and Syst.}, Madrid, Spain, Oct
  2018, pp. 1672--1679.

\bibitem{kozak_ranger_2016}
K.~Kozak and M.~Alban, ``Ranger: A ground-facing camera-based localization
  system for ground vehicles,'' in \emph{Proc. IEEE/ION Position, Location and
  Navigation Symp.}, Savannah, GA, USA, Apr 2016, pp. 170--178.

\bibitem{shan_lvi-sam_2021}
T.~Shan, B.~Englot, C.~Ratti, and D.~Rus, ``{LVI-SAM}: Tightly-coupled
  lidar-visual-inertial odometry via smoothing and mapping,'' in \emph{Proc.
  IEEE Int. Conf. Robot. and Automat.}, Xi'an, China, May 2021, pp. 5692--5698.

\bibitem{cheng_visual-laser-inertial_2021}
D.~Cheng, H.~Shi, A.~Xu, M.~Schwerin, M.~Crivella, L.~Li, and H.~Choset,
  ``Visual-laser-inertial {SLAM} using a compact {3D} scanner for confined
  space,'' in \emph{Proc. IEEE Int. Conf. Robot. and Automat.}, Xi'an, China,
  May 2021, pp. 2450--2456.

\bibitem{zubizarreta_direct_2020}
J.~Zubizarreta, I.~Aguinaga, and J.~Montiel, ``Direct sparse mapping,''
  \emph{IEEE Trans. on Robot.}, vol.~36, no.~4, pp. 1363--1370, May 2020.

\bibitem{rublee_orb_2011}
E.~Rublee, V.~Rabaud, K.~Konolige, and G.~Bradski, ``{ORB}: an efficient
  alternative to {SIFT} or {SURF},'' in \emph{Proc. IEEE Int. Conf. Comput.
  Vision}, Barcelona, Spain, Nov 2011, pp. 2564--2571.

\bibitem{lowe_distinctive_2004}
D.~G. Lowe, ``Distinctive image features from scale-invariant keypoints,''
  \emph{Int. J. of Comput. Vision}, vol.~60, no.~2, pp. 91--110, Nov 2004.

\bibitem{zhang_front-end_2021}
Y.~Zhang and J.~Leonard, ``A front-end for dense monocular {SLAM} using a
  learned outlier mask prior,'' in \emph{Proc. IEEE Int. Conf. Robot. and
  Automat.}, Xi'an, China, May 2021, pp. 11\,732--11\,738.

\bibitem{mur-artal_orb-slam_2015}
R.~Mur-Artal, J.~M.~M. Montiel, and J.~D. Tardos, ``{ORB-SLAM}: A versatile and
  accurate monocular {SLAM} system,'' \emph{IEEE Trans. Robot.}, vol.~31,
  no.~5, pp. 1147--1163, Oct 2015.

\bibitem{mur-artal_orb-slam2_2017}
R.~Mur-Artal and J.~D. Tardos, ``{ORB-SLAM2}: An open-source {SLAM} system for
  monocular, stereo, and {RGB-D} cameras,'' \emph{IEEE Trans. Robot.}, vol.~33,
  no.~5, pp. 1255--1262, Oct 2017.

\bibitem{campos_orb-slam3_2021}
C.~Campos, R.~Elvira, J.~J.~G. Rodriguez, J.~M. M.~Montiel, and J.~D.~Tardos,
  ``{ORB-SLAM3}: An accurate open-source library for visual, visual–inertial,
  and multimap {SLAM},'' \emph{IEEE Trans. Robot.}, vol.~37, no.~6, pp.
  1874--1890, Dec 2021.

\bibitem{quan_monocular_2021}
M.~Quan, S.~Piao, Y.~He, X.~Liu, and M.~Qadir, ``Monocular visual {SLAM} with
  points and lines for ground robots in particular scenes: parameterization for
  lines on ground,'' \emph{J. of Intell. and Robotic Syst.: Theory and Appl.},
  vol. 101, no.~4, Mar 2021.

\bibitem{pumarola_pl-slam_2017}
A.~Pumarola, A.~Vakhitov, A.~Agudo, A.~Sanfeliu, and F.~Moreno-Noguer,
  ``{PL-SLAM}: real-time monocular visual {SLAM} with points and lines,'' in
  \emph{Proc. IEEE Int. Conf. Robot. and Automat.}, Singapore, May 2017, pp.
  4503--4508.

\bibitem{civera_inverse_2008}
J.~Civera, A.~Davison, and J.~Montiel, ``Inverse depth parametrization for
  monocular {SLAM},'' \emph{IEEE Trans. Robot.}, vol.~24, no.~5, pp. 932--945,
  Oct 2008.

\bibitem{laina_deeper_2016}
I.~Laina, C.~Rupprecht, V.~Belagiannis, F.~Tombari, and N.~Navab, ``Deeper
  depth prediction with fully convolutional residual networks,'' in \emph{Proc.
  IEEE Int. Conf. 3D Vision}, Stanford, California, USA, Oct 2016, pp.
  239--248.

\bibitem{kim_revisiting_2021}
U.-H. Kim, G.-M. Lee, and J.-H. Kim, ``Revisiting self-supervised monocular
  depth estimation,'' in \emph{Robot Intell. Technol. and Appl. 6 - Results 9th
  Int. Conf. Robot Intell. Technol. and Appl.}, Mar 2021, pp. 336--350.

\bibitem{zhang_high-precision_2019}
L.~Zhang, A.~Finkelstein, and S.~Rusinkiewicz, ``High-precision localization
  using ground texture,'' in \emph{Proc. IEEE Int. Conf. Robot. Automat.},
  Montreal, QC, Canada, May 2019, pp. 6381--6387.

\bibitem{fabian_schmid_ground_2020}
J.~F. Schmid, S.~F. Simon, and R.~Mester, ``Ground texture based localization:
  do we need to detect keypoints?'' in \emph{Proc. IEEE/RSJ Int. Conf. Intell.
  Robots and Syst.}, Las Vegas, NV, USA, Oct 2020, pp. 4575--4580.

\bibitem{zhang_learning_2018}
L.~Zhang and S.~Rusinkiewicz, ``Learning to detect features in texture
  images,'' in \emph{Proc. IEEE/CVF Conf. Comput. Vision and Pattern
  Recognit.}, Salt Lake City, UT, USA, Jun 2018, pp. 6325--6333.

\bibitem{muja_fast_2009}
M.~Muja and D.~Lowe, ``Fast approximate nearest neighbors with automatic
  algorithm configuration,'' in \emph{Proc. 4th Int. Conf. Comput. Vision
  Theory and Appl.}, Lisboa, Portugal, Feb 2009, pp. 331--340.

\bibitem{bradski_opencv_2000}
G.~Bradski, ``The {OpenCV} library,'' \emph{Dr. Dobb's Journal of Software
  Tools}, 2000.

\bibitem{dellaert_borglabgtsam_2021}
F.~Dellaert, ``Factor graphs and {GTSAM}: a hands-on introduction,'' Georgia
  Institute of Technology, Atlanta, GA, USA, Tech. Rep. GT-RIMCP\&R-2012-002,
  Sep 2012.

\bibitem{galvez-lopez_bags_2012}
D.~Galvez-López and J.~D. Tardos, ``Bags of binary words for fast place
  recognition in image sequences,'' \emph{IEEE Trans. on Robot.}, vol.~28,
  no.~5, pp. 1188--1197, Oct 2012.

\bibitem{schmid_hd_2022}
J.~F. Schmid, S.~F. Simon, R.~Radhakrishnan, S.~Frintrop, and R.~Mester, ``{HD}
  ground - a database for ground texture based localization,'' in \emph{Proc.
  IEEE Int. Conf. Robot. and Automat.}, Philadelphia, PA, USA, May 2022, pp.
  7628--7634.

\end{thebibliography}
\end{document}